\documentclass[a4paper,conference]{IEEEtran}
\IEEEoverridecommandlockouts
\usepackage{cite}
\usepackage{amsmath,amssymb,amsfonts}
\usepackage{algorithmic}
\usepackage{graphicx}
\usepackage{textcomp}
\usepackage{xcolor}
\def\BibTeX{{\rm B\kern-.05em{\sc i\kern-.025em b}\kern-.08em
    T\kern-.1667em\lower.7ex\hbox{E}\kern-.125emX}}
    
\usepackage{multirow}
    
\usepackage{amsfonts}

\usepackage{tabularx}
    \newcolumntype{L}{>{\raggedright\arraybackslash}X}
    \newcolumntype{C}{>{\raggedright\arraybackslash}Z}
\newcommand{\heading}[1]{\multicolumn{1}{|c|}{#1}}

\usepackage{siunitx}

\usepackage{algorithm}

\begin{document}

\title{Adversarial Encoder-Multi-Task-Decoder for Multi-Stage Processes}

\author{\IEEEauthorblockN{Andre Mendes}
\IEEEauthorblockA{\textit{New York University} \\
New York, USA \\
andre.mendes@nyu.edu}
\and
\IEEEauthorblockN{Julian Togelius}
\IEEEauthorblockA{\textit{New York University} \\
New York, USA \\
julian.togelius@nyu.edu}
\and
\IEEEauthorblockN{Leandro dos Santos Coelho}
\IEEEauthorblockA{\textit{Pontifical Catholic University of Parana} \\
\textit{Federal University of Parana}\\
Curitiba, Brazil \\
leandro.coelho@pucpr.br}
}

\maketitle

\begin{abstract}
In multi-stage processes, decisions occur in an ordered sequence of stages. Early stages usually have more observations with general information (easier/cheaper to collect), while later stages have fewer observations but more specific data. This situation can be represented by a dual funnel structure, in which the sample size decreases from one stage to the other while the information increases. Training classifiers in this scenario is challenging since information in the early stages may not contain distinct patterns to learn (underfitting). In contrast, the small sample size in later stages can cause overfitting. We address both cases by introducing a framework that combines adversarial autoencoders (AAE), multi-task learning (MTL), and multi-label semi-supervised learning (MLSSL). We improve the decoder of the AAE with an MTL component so it can jointly reconstruct the original input and use feature nets to predict the features for the next stages. We also introduce a sequence constraint in the output of an MLSSL classifier to guarantee the sequential pattern in the predictions. Using real-world data from different domains (selection process, medical diagnosis), we show that our approach outperforms other state-of-the-art methods.
\end{abstract}

\section{Introduction}
In many real-world applications, decisions are taken following a sequence of steps or stages. This type of process can be defined as a multi-stage process, and they have an important characteristic, which is the relationship between the information available and sample size in each stage. Initial stages have a large sample size and general/cheap information, whereas, in later stages, more data is collected but for a smaller selected group, which reduces the sample size. 

For medical diagnosis~\cite{caruana1997multitask}, for example, in initial stages, low-cost information such as demographics (city, state) and physical attributes (weight, height, BMI) are collected for a larger population. From this information, a group is selected to perform more expensive tests (MRI, glucose, ECG). Therefore, the sample size decreases (selected group) while the information increases (more tests).

Another example is the hiring process. Applicants submit general information in the initial stages, such as resumes, and an evaluator screens trough them to select people to move on to the next round. This process continues until the final pool is selected. In terms of information, initial stages have general data about the applicants, whereas more information is gathered as the process goes on. Hence, in the final stages, the evaluator has a smaller pool but with much more information about each applicant.

\begin{figure}[tbp]
\centerline{\includegraphics[width=1\linewidth]{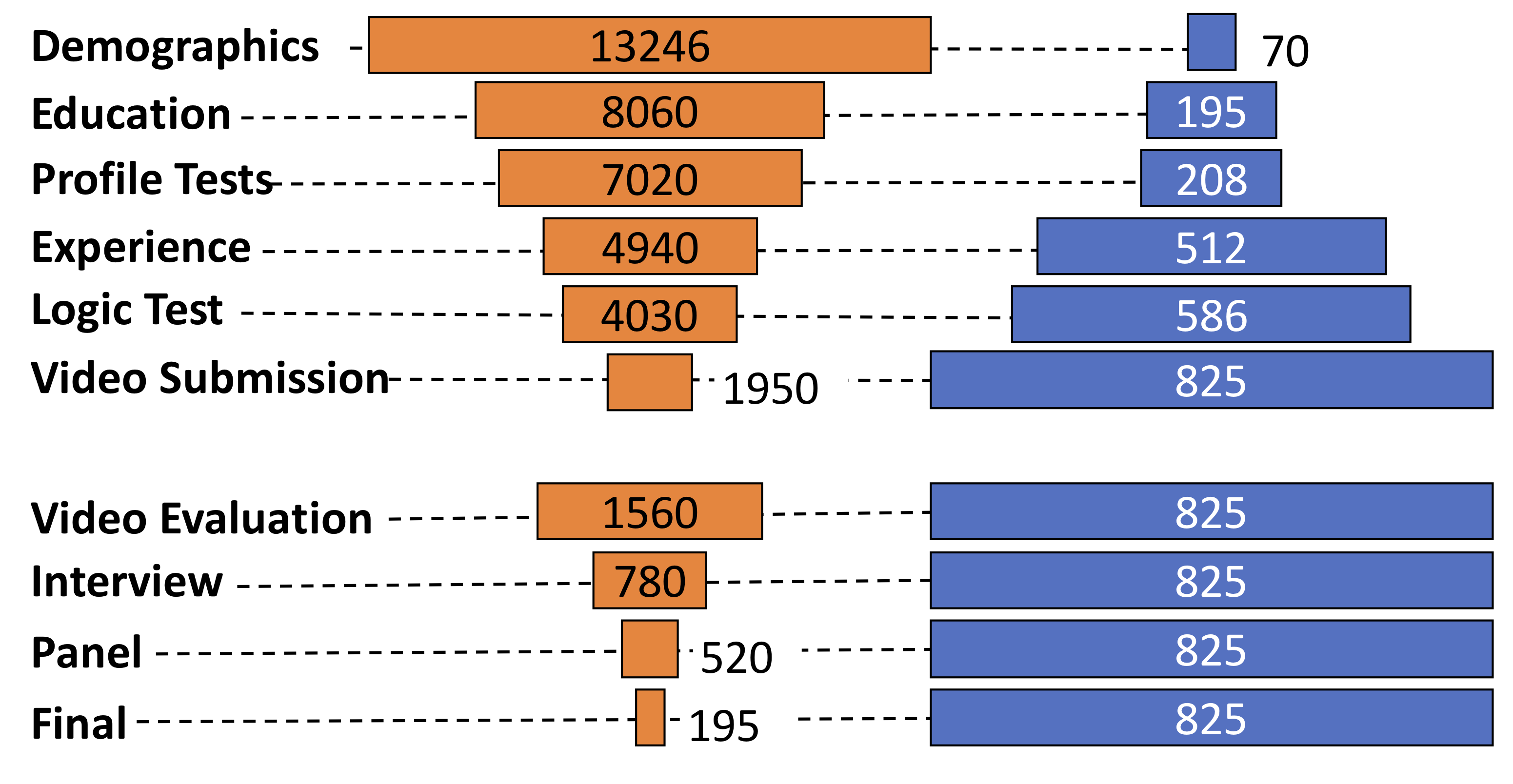}}
\caption{Representation of a dual funnel problem. The left funnel shows the number of applicants decreasing, whereas the right funnel shows the amount of data (in terms of variables) increasing during the process.}
\label{fig:dual_funnel}
\end{figure}

Learning classifiers for multi-stage processes can be hard due to the dual funnel structure, as shown in Fig.~\ref{fig:dual_funnel}. During the initial stages, the dataset grows in dimensionality but decreases in terms of sample size. Classifiers trained in these stages have sufficient samples to generalize, but the available features might not contain enough valuable information, causing high bias and underfitting. In the final stages, there is more distinct, richer information, but the sample size is reduced significantly, causing classifiers to suffer from high variance and overfitting. To address this problem, we redesign the multi-stage structure and present a framework with two components.

The first component is based on an \textit{adversarial autoencoder} (AAE)~\cite{makhzani2015adversarial}. Traditionally, the encoder and decoder in an AAE are similar neural network structures used to learn latent features and reproduce the input dataset. In this work, we improve the decoder using multi-task learning (MTL)~\cite{caruana1997multitask}, so it can jointly reconstruct the original input and predict the features for the next stages. We refer to the predictions of future features as feature nets, inspired on ~\cite{caruana1997multitask}. We assume that current and future features are correlated, and in some cases, dependent on each other, and these relationships can help the decoder to learn the entire dataset better. The final structure is called adversarial encoder-multi-task decoder (AEMTD), and it used to generate a complete dataset from any stage.

In the second component, we redefine the multi-stage problem from binary to multi-label (ML) classifications, using label relationships as additional information. To consider the sequence structure of the problem, we also add a sequence constraint on the predictions. Additionally, to use unlabelled generated samples from AEMTD, we adopt a semi-supervised learning (SSL) approach to enforce label consistency between similar input features. Finally, our second component is a multi-label semi-supervised learning (MLSSL) classifier. 

We refer to the entire framework as STAGE, and the main contributions of this paper are:
\begin{enumerate}
    \item We combine adversarial autoencoders with feature nets, creating a multi-task decoder to generate a complete dataset to be shared in different stages.
    \item We define the multi-stage problem as multi-label to force a classifier to learn correlation among labels in different stages. We also propose a sequence constraint that penalizes predictions nonconforming with the sequential structure of the process.
    \item We combine ML and SSL to use labeled and unlabeled generated data during training.
    \item The effectiveness of our framework is demonstrated by extensive experiments on real data from distinct domains. Our model outperforms other state-of-art frameworks, particularly in the later stages, where the sample size is significantly reduced.
\end{enumerate}

The rest of this paper is organized as follows: Section~\ref{sec:related} shows the work related to this approach; in Section~\ref{sec:problem}, we present the problem definition and in Section~\ref{sec:methods} we propose our method. We describe the experiments and results for validation in Section~\ref{sec:experiments} and Section~\ref{sec:results}, respectively. Finally, the present a conclusion in Section~\ref{sec:conclusion}.
\section{Related Work}
\label{sec:related}
The two components in our framework are built based on adversarial autoencoder (AAE)~\cite{baldi2012autoencoders,makhzani2015adversarial} and multi-task learning (MTL)~\cite{zhang2017survey,caruana1997multitask}; as well as multi-label classification (MLC)~\cite{zhang2013review} and semi-supervised learning (SSL)~\cite{zhu2005semi}; respectively. Additionally, our work is also related to multi-stage classifiers~\cite{trapeznikov2012multi,viola2001robust}. Here we describe the work related to ours and refer the reader to the above citations for a deeper understanding of each field.
\subsection{Autoencoders and Multi-Task Learning} 
Autoencoders are learning models that transform inputs into outputs with the least possible amount of distortion~\cite{rumelhart1985learning,erhan2010does}. Recently, deep models have a focus on enriching the prior or posterior of explicit generative models such as variational autoencoders (VAEs)~\cite{kingma2013auto,rezende2015variational} or use GANs~\cite{goodfellow2014generative} for alternative training objectives to the log-likelihood~\cite{arjovsky2017wasserstein,zhu2017unpaired}. The AAE in~\cite{makhzani2015adversarial} uses an adversarial approach so that the encoder learns to convert the data distribution to the prior distribution. At the same time, the decoder maps the imposed prior to the data distribution.

In MTL, multiple related tasks are learned simultaneously so that knowledge can be shared among them. Different approaches for MTL include neural nets (NN) and kernel methods~\cite{caruana1997multitask,kumar2012learning} as well as deep neural nets (DNN)~\cite{ruder2017overview}. The regularization parameters in MTL control how information is shared between tasks and prevents overfitting~\cite{kumar2012learning}.

The combination of MTL and AAE has been proposed in different fields. For speech synthesis, the architecture in~\cite{yang2017statistical} combines traditional acoustic loss function and the GAN's discriminative loss. In~\cite{liu2018multi}, an MTL encoder-discriminator-generator is presented for disentangled feature learning.
\subsection{Multi-Label and Semi-Supervised Learning}
In MLC, one instance can be assigned to several categories simultaneously~\cite{zhang2013review}. Methods designing MLC as multiple binary tasks~\cite{tsoumakas2007multi} lose information on the correlation between labels. To overcome this, in ~\cite{read2011classifier}, chain classifiers are used to explore cross-label prior information. In Label Embeddings, labels are mapped into a subspace with latent embedding so that the correlation between them can be implicitly used~\cite{tai2012multilabel,chen2012feature,wang2016cnn}.

SSL is applied when a dataset contains labeled and unlabeled observations. In Graph-based methods, the goal is to construct a graph connecting similar observations, so that label information propagates through the graph~\cite{blum2004semi}. In NN methods, some approaches combine SSL with generative models~\cite{kingma2014semi} to employ rich parametric density estimators, or virtual adversary training~\cite{miyato2018virtual} for better regularization using adversarial direction without label information.

Some methods tackle MLC and SSL (MLSSL) by using a transductive setting with label smoothness regularization~\cite{wu2014multi} or by formulating a convex quadratic matrix optimization problem~\cite{wu2018multi}. Furthermore, the Canonical-Correlated Autoencoder~\cite{yeh2017learning} performs feature and label embedding jointly with label prediction in an end-to-end process.   
\subsection{Multi-Stage Classification}
Similar methods have been proposed for multi-stage classification. Cascades classifiers~\cite{viola2001robust} make partial decisions, delaying a positive decision until the final stage. In contrast, multi-stage classifiers~\cite{trapeznikov2012multi} can deal with multi-class problems and can make classification decisions at any stage. In~\cite{sabokrou2017deep}, a deep multi-stage approach creates classifiers that are jointly optimized and cooperate across stages.
\section{Problem Definition}
\label{sec:problem}
In a multi-stage process with $S$ stages, we define a single stage as $s$, with $s=\{0,1,...,S\}$. Every stage has a dataset with training samples $\{x_i^s,y_i^s\}_{i=0}^{n^s}$, where $n^s$ refers to the number of samples. The number of features is given by $d^s$, so the feature input is given by $X^s \in \mathbb{R}^{n^s\times d^s}$ and the labels by $Y^s \in \mathbb{R}^{n^s \times 1}$. Let's also define $A \in \mathbb{R}^{n}$ as the vector of all applicants, where $n$ is given by $n=\sum_{s=1}^S n^s$ and $a \in A$ represents a single applicant. The feature vector for a single applicant $a$ in stage $s$ is given by the vector $x_a^s \in \mathbb{R}^{d^s}$. A prediction matrix $\hat{Z} \in \mathbb{R}^{n \times S}$ can be defined, where $z_{a} \in \mathbb{R}^{S}$ is the prediction vector for an applicant $a$ in all stages, and $z^s_{a} \in \{-1,1\}$ represents the prediction for a single stage $s$. If the predictions in this matrix are reliable, it is possible to estimate the chances of an applicant in the entire process.

\subsection{Underfitting in Earlier Stages}
 In each stage $s$, only features $x^s$ up to that stage are available. Therefore, a prediction for an applicant with data up to $s$ is given by $z_a^i=f(x^j_a)$, where $j=\{0,...,s\}$, $i=\{0,...,S\}$ and $f(\cdot)$ is a classification function. For example, for an applicant in stage $s=0$, all predictions will be made using $x_a^0$. 

Since the features in early stages are general and less discriminative, models trained in these stages have poor performance predicting the applicant's future in the process. To address this problem, one could incorporate future features in the early stages. More specifically, if we can reliably create a data prediction function $g(\cdot)$ to predict the features $\hat{\text{f}}$ for later stages, we can create a complete dataset for all stages.

By combining $x^s$ and $\hat{\text{f}}$, we obtain the feature vector $\hat{x}$. Applying this process to all applicants, we create the complete dataset $\hat{X} \in \mathbb{R}^{n \times {d^S}}$, which contains more discriminative information to make better predictions in early stages.
\subsection{Overfitting in Later Stages}
For each new stage, new data is received while the number of samples decrease, which means $X^{s+1}\neq X^{s}$, $n^{s+1}<n^{s}$ and $d^{s+1}>d^{s}$.  As $n^s$ gets significantly smaller in absolute value and in comparison to $d^s$, classifiers trained on the specific stage data tend to overfit. One possible way to address this problem is to use the predicted complete dataset $\hat{X}$.

This results in more training samples that can be used to train classifiers in later stages. However, $\hat{X}=g(X^s)$ only generate new samples but no labels, since the applicants in $s$ were not evaluated in stages posterior to $s$. Hence, we only have $n^s$ labels, while the complete dataset has $n$ samples. In this case, SSL can be applied, so that labeled and unlabeled data are combined to create a better classifier. 
\subsection{Tasks Relationships and Sequence Constraints}
With $S$ stages, we could construct $S$ classifiers to predict the outcomes in each stage. However, since we can generate the complete dataset, we can also train one classifier using the relationship between tasks to make predictions for all stages together. Hence, we redefine the problem from multiple binary classifications to a single multi-label classification problem.

Additionally, we also consider the sequence of the process. For example, an applicant approved in stage $s$ can not be rejected in a previous stage $s-1$. Mapping these relationships adds sequence information to the classifier, enforcing predictions to be consistent between previous and posterior stages.
\section{Methods}
\label{sec:methods}
\begin{figure*}[tbp]
\centerline{\includegraphics[width=1\linewidth]{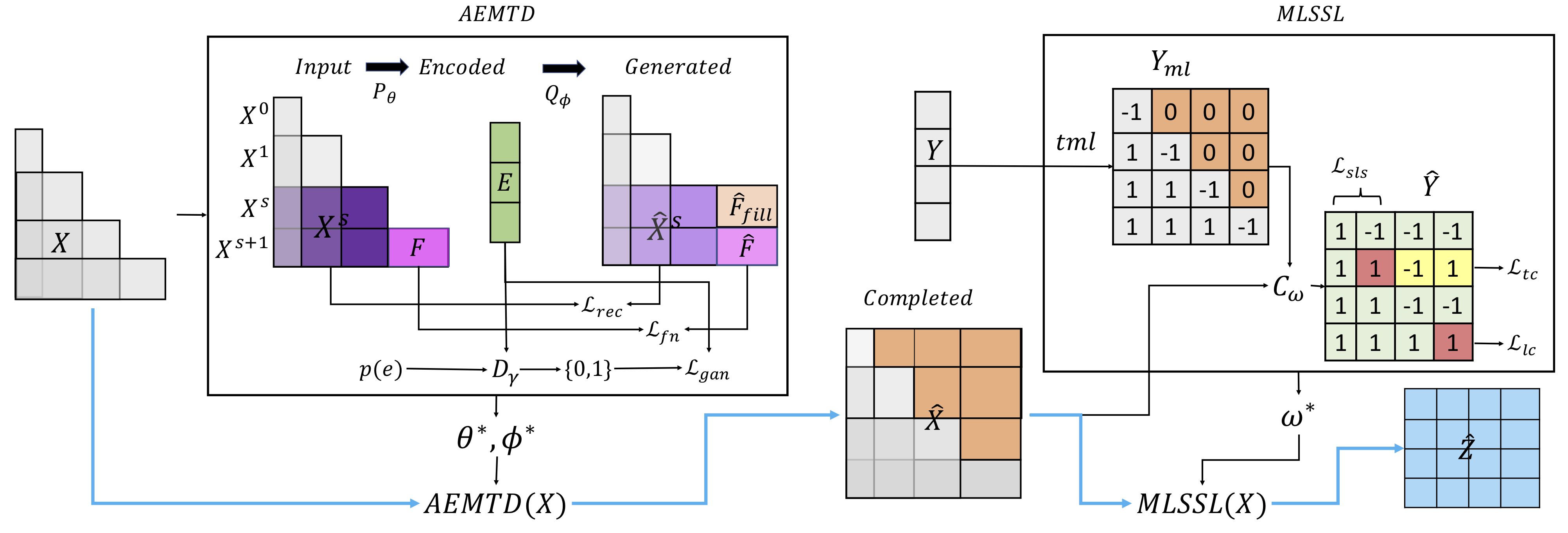}}
\caption{Representation of the STAGE framework. On the left, a dataset $X$ with original data for the all stages is fed to the AEMTD component. The encoder $P_\theta$ produces encoded values $E$ that can't be distinguished from the prior distribution $p(e)$ by the discriminator $D_\gamma$. The decoder $Q_\phi$ uses $E$ to reproduce an input $X^S$ and to predict future features $F$. After training is complete, optimal parameters $\theta^*$ and $\phi^*$ are obtained. Using these parameters, a function $AEMTD(\cdot)$ generates a complete dataset $\hat{X}$ from $X$. On the right side, the multi-class labels $Y$ are converted to multi-label values $Y_{ml}$ with missing labels filled with $0$. $\hat{X}$ and $Y_{ml}$ are used to train the classifier $C_\omega$ using a MLSSL approach. $C_\omega$ is penalized for label inconsistency ($\mathcal{L}_{lc}$ - red), sample level consistency ($\mathcal{L}_{sls}$) and sequence consistency ($\mathcal{L}_{tc}$ - yellow). When training is done, optimal parameters $\omega^*$ are obtained. On the bottom, blues lines indicate the workflow in production. Given $X$, $AEMTD(X)$ generates $\hat{X}$ and $MLSSL(\hat{X})$ creates the prediction matrix $\hat{Z}$.}
\label{fig:aemtd_mlssl}
\end{figure*}

In this section we explain all the components for our framework shown in Fig.~\ref{fig:aemtd_mlssl}.
\subsection{Adversarial Autoencoder (AAE)}
The first component in our method is an AAE~\cite{makhzani2015adversarial}, which has an encoder $P_\theta$, with parameters $\theta$, and a decoder $Q_\phi$ with parameters $\phi$. Considering an stage $s$, the goal of the encoder is to map an input feature vector $x^s \in \mathbb{R}^{d^s}$ to an embedding space $e \in \mathbb{R}^{d_e}$. The decoder is used to reproduce the input feature vector, $\hat{x}^s$ from the embedding space $e$. Using the original and predicted input, the reconstruction error is
\begin{align}
\label{eq:rec_loss}
 \mathcal{L}_{rec}= \sum^{n^s}_{i=1}\sum^{d^s}_{j=1}L_f(\hat{x}_{ij},x_{ij}),
\end{align}
where
\[
 L_f(x_{ij},\hat{x}_{ij})=
 \begin{cases}
(\hat{x}_{ij}-x_{ij})^{2}, & \text{if \ensuremath{x_{ij}} is continuous},\\
-x_{ij}\log(\hat{x}_{ij}), & \text{if \ensuremath{x_{ij}} is binary}.
\end{cases}
\]

To regularize the autoencoder, an adversarial network is attached on top of its embedding space $e$. Such network is used to match the aggregated posterior, $P_\theta(e)$, to an arbitrary prior, $p(e)$. Therefore, the generator (the encoder $P_\theta$) ensures the aggregated posterior distribution can fool the discriminator $D_\gamma$ with parameters $\gamma$. In other words, $D$ can't distinguish if the embedding $e$ comes from the true prior distribution $p(e)$. This adversarial approach is expressed as
\begin{equation}
\label{eq:minmax}
    \min_G \min_D E_{x\sim p_{d}}[\log D(x)] + E_{e\sim p(e)}[log(1-D(P_\theta(e))].
\end{equation}

We refer to Eq.~\ref{eq:minmax} as $\mathcal{L}_{gan}$ and $p_d$ is the distribution of the data. We choose a Gaussian distribution for $p(e)$ and use the encoder to predict its mean and variance. 
\subsection{Multi-Task Decoder with Feature Nets }
In addition to predicting the input feature, we expand the decoder by adding a multi-task component creating the Adversarial Encoder-Multi-Task-Decoder structure (AEMTD). Our goal is to reconstruct the input and also predict the features from a future stage. We call this approach feature nets, which is inspired by~\cite{caruana1997multitask}. We assume that current and future features are correlated, and the decoder can benefit from learning them together. 

For a given stage $s$, the matrix features can be expressed as $X_{i=n-n^s:n \ ; \ j=0:d^s}$. The difference from features in a stage $s$ and a stage $s+1$ is given by $F=X_{i=n-n^{s+1}:n \ ; \  j=d^s:d^S}$. The features $\text{F}$ (Shown in magenta in Fig.~\ref{fig:aemtd_mlssl}) are the ones we aim to predict using the decoder, hence $\hat{\text{F}}=Q_\phi(X^s)$. The loss for the feature nets part is given by
\begin{align}
\label{eq:features_net_loss}
 \mathcal{L}_{fn}= \sum^{n}_{i=n-n^{s+1}}\sum^{d^S}_{j=d^s}L_f(\hat{x}_{ij},x_{ij}).
\end{align}

The final loss for the AEMTD component is
\begin{equation}
\label{eq:aemtd_loss}
    \arg \min_{P_\theta,Q_\phi} \max_{D_\gamma} \mathcal{L}_{rec}+
    \mathcal{L}_{fn}+
    \mathcal{L}_{gan}.
\end{equation}

Similar to the original AAE, all the parts of the AEMTD can be trained jointly using Stochastic Gradient Descent (SGD) in two phases,  reconstruction and regularization. 

In the reconstruction, the encoder and the decoder are updated to minimize the reconstruction error and the prediction of future features. In the regularization phase, the discriminator is updated to differentiate true and generated samples. Finally, the generator is updated to confuse the discriminator. Once the training is done, the decoder defines a generative model that maps the imposed prior of $p(e)$ to the data distribution.
\subsection{Constrained Multi-Label Learning with Incomplete Labels}
With the trained AEMTD, we can generate $\hat{X}^s$, and the predicted future features, $\hat{\text{F}}$. By combining them, we create the complete dataset $\hat{X}=\hat{X}^s\oplus \hat{\text{F}}$, where $\oplus$ is the completing operation shown in orange in Fig.~\ref{fig:aemtd_mlssl}. 

Instead of creating individual classifiers for each stage and having multiple binary classifications, we redefine the problem as multi-label classification. We combine the labels from each stage into a label vector $y_a \in \{1,-1\}^{S}$, where $y_a^s \in \{-1,1\}$ represents the result for an applicant $a$ in stage $s$.

When the complete dataset is created using the decoder $Q_\theta$, only samples are generated. As a result, the applicants that were not evaluated in advanced stages don't have the respective labels. Therefore, we represent approval, rejection, and missing labels as 1, -1, and 0, respectively. Finally, we use available and missing labels during training, similar to~\cite{wu2014multi}. 

Given $Y \in \{1,0,-1\}^{n \times S}$, our goal is to predict a complete label matrix $\hat{Y} \in \{1,-1\}^{n \times S}$ from $Y$ using three constraints:
\begin{itemize}
    \item Label Consistency - The predicted label matrix $\hat{Y}$ should be consistent with the initial label matrix $Y$;
    \item Label Smoothness -  If the two samples $x_i$ and $x_j$ are similar, then their labels, i.e., the corresponding column vectors of $\hat{Y}$ should be similar
    \item Sequence Consistency - We want to make the predictions consistent with the sequence of stages, which is defined using two rules: (1) if a label value for stage $s$ is 1, the label values for all stages before $s$ also have to be 1; (2) if a label value for stage $s$ is -1, the label values for all stages after $s$ have to be -1. This is similar to the idea of thermometer encoding.
\end{itemize}

For label consistency, the loss is given by
\begin{equation}
    \mathcal{L}_{lc} = ||\hat{Y}-Y||^2_\mathcal{F},
\end{equation}
where $||\cdot||_\mathcal{F}$ indicates the \textit{Frobenius} norm. 

For sample smoothness, we first define a sample similarity matrix with all pairwise correlations among $\hat{X}$ using
\begin{equation}
    V_X(i,j)=\exp(\frac{-m^2(x_i,x_j)}{\sigma_i\sigma_j}),
\end{equation}
which is based on a k-nn graph so that $V_{ij}=0$ if $x_j$ is not within the $k_{nn}$-nearest neighbors of $x_i$ and $V_X(i,i)=0$. We use the Euclidean distance for $m(x_i,x_j)$ and $\sigma_i=d(x_i,x_h)$, with $x_h$ as the $h_{nn}$-th nearest neighbor of $x_i$.

We use a normalization term defined as $d_X(i)=\sum_j^n V_X(i,j)$~\cite{von2007tutorial} to make the smoothness term invariant to the different scaling factors of the elements of $V_X$. Using this term, we can define the matrix $D_X=diag(d_X(1),...,d_X(n))$. The final sample level smoothness term is given by:
\begin{equation}
    \mathcal{L}_{sls} = tr(\hat{Y}L_X\hat{Y}^T),
\end{equation}
where $L_X=I-D_X^{-\frac{1}{2}}V_XD_X^{-\frac{1}{2}}$ and $tr$ is the \textit{trace} of a matrix. 


For sequential consistency, we observe that our 2 rules can be simplified by verifying the occurrence of the pair $(-1,1)$ in any location in the output multi-label string. This pair break both of the defined sequential rules, and the model is penalized when such predictions are made. Therefore, we use the following loss to perform sequential consistency
\begin{equation}
    \mathcal{L}_{tc}=\frac{1}{4}\sum_{a=1}^n \sum_{i<j}(1-\hat{y}_a^i)(1+\hat{y}_a^j).
\end{equation}

Finally, we create the MLSSL classifier $C_\omega$ by solving
\begin{equation}
\label{eq:mlssl_loss}
    \arg \min_{C_{\omega}} \mathcal{L}_{lc}+
    \mathcal{L}_{sls}+
    \mathcal{L}_{tc}+
    \lambda||\omega||_F^2,
\end{equation}
where $\omega$ are the parameters for the classifier and the hyperparameter $\lambda$ controls the \textit{l2}-norm penalty to prevent overfitting.


\subsection{Training Procedure}
\label{subsec:training_procedure}
To train the entire framework (see Fig.~\ref{fig:aemtd_mlssl}), we use two procedures that go backwards in terms of the order of stages.

For the AEMTD component (see Alg.~\ref{alg:aemtd}), we start from the last stage $S$. Since the input dataset is complete, this step can be considered as initializing the AEMTD with all features. We use Equations~\ref{eq:rec_loss} and~\ref{eq:minmax} to update parameters $\theta$ and $\phi$ for the encoder-decoder and parameters $\gamma$ for the discriminator. When training using data from this stage is complete, the decode $Q_\phi$ can generate a complete dataset $\hat{X} \in \mathbb{R}^{n \times S}$. We continue the training backwards from $s=S-1$ to $s=0$, however, we now add the feature nets loss in Eq.~\ref{eq:features_net_loss} to train the AEMTD and generate the complete dataset.

For the MLSSL component (see Alg.~\ref{alg:mlssl}), in each stage, the complete dataset is created using $\hat{X}\leftarrow AEMTD(X^s)$. We also transform the labels from binary to ML, hence $Y_{ml}=tml(Y)$. For $s=S$, the complete labels $Y_{ml} \in \{1,-1\}^{n \times S}$ are used and from $s=S-1$ to $s=0$, we add the missing labels to create  $Y_{ml} \in \{1,0,-1\}^{n \times S}$. In all steps during training, we use Eq.~\ref{eq:mlssl_loss} to update the classifier parameters $\omega$.

By the end of the process, given data from a stage, AEMTD ($P_\theta,Q_\phi$) generates a complete dataset, and MLSSL ($C_\omega$) makes predictions for all stages.

\begin{algorithm}[tb]
\caption{Train AEMTD}
\label{alg:aemtd}
\textbf{Input}: $X,k_{mb},p(e)$, initialized $P_\theta,Q_\phi,D_\gamma$\\
\textbf{Output}: Optimized $P_{\theta^*},Q_{\phi^*}$
\begin{algorithmic}[] 
    \WHILE{stop criterion not met}
        \FOR{$s=S$ to 0}
            \STATE \textbf{Discriminator Optimization}
            \STATE Get  $X^s_{mb}$ with $k_{mb}$ random samples from $X^s$
            \STATE $E\leftarrow P\theta(X^s_{mb})$ and get $k_{mb}$ samples from $p(e)$
            \STATE Compute the gradient in Eq.~\ref{eq:aemtd_loss} w.r.t $D_\gamma$
            \STATE Take a step to update $D_\gamma$ to maximize Eq.~\ref{eq:aemtd_loss}
            \STATE \textbf{Encoder-Multi-Task-Decoder Optimization}
            \STATE Get  $X^s_{mb}$ with $k_{mb}$ random samples from $X^s$
            \STATE $E\leftarrow P_\theta(X^s_{mb})$ and $\hat{X}^s_{mb},\hat{F}\leftarrow Q_\phi(E)$
            \STATE Compute the gradient in Eq.~\ref{eq:aemtd_loss} w.r.t $P_\theta$, $Q_\phi$
            \STATE Take a step to update $P_\theta$, $Q_\phi$ to minimize Eq.~\ref{eq:aemtd_loss}
        \ENDFOR
    \ENDWHILE
\end{algorithmic}
\end{algorithm}

\begin{algorithm}[tb]
\caption{Train MLSSL}
\label{alg:mlssl}
\textbf{Input}: $X,Y,k_{mb},\lambda, h_{nn},k_{nn}$, initialized $C_\omega$\\
\textbf{Output}: Optimized $C_{\omega^*}$
\begin{algorithmic}[] 
    \WHILE{stop criterion not met}
        \FOR{$s=S$ to 0}
            \STATE Get  $X^s_{mb}$ with $k_{mb}$ random samples from $X^s$
            \STATE $\hat{X}_{mb}\leftarrow AEMTD(X^s_{mb})$ and $Y_{ml}\leftarrow tml(Y)$
            \STATE $\hat{Y}\leftarrow C_\omega(\hat{X})$
            \STATE Compute the gradient in Eq.~\ref{eq:mlssl_loss} w.r.t $C_\omega$
            \STATE Take a step to update $C_\omega$ to minimize Eq.~\ref{eq:mlssl_loss}
        \ENDFOR
    \ENDWHILE
\end{algorithmic}
\end{algorithm}

\section{Experiments}
\label{sec:experiments}
Here we show the application of our method in two real-world domains: selection process and medical diagnosis. 
\subsection{Selection Process}
For the selection process, we use datasets from 2 different companies with similar recruitment processes. We refer to them with indexes such that $C_1$ indicates Company 1. Although their processes and target group are similar (senior undergraduate students), their goals are different. $C_1$ is an organization that selects students for a fellowship, while $C_2$ is a retail company that focuses on its recent-grad hire program.

In both companies, the processes happen annually, and we have data for three years. The dual funnel structure for $C_2$ is shown in Fig.~\ref{fig:dual_funnel}. The process in $C_1$ has the same structure, and, on average, 35 out of 19000 applicants are selected.
\begin{table*}[tbp]
\caption{Stages in the multi-stage selection process}
\label{tab:process_description}
\scriptsize
\begin{tabularx}{\linewidth}{|>{\hsize=.11\hsize\linewidth=\hsize}X|>{\hsize=.52\hsize\linewidth=\hsize}X|>{\hsize=.35\hsize\linewidth=\hsize}X|}
\hline
\heading{Stages} & \heading{Company 1 ($C_1$)} & \heading{Company 2 ($C_2$)} \\ \hline
Demographics & Provide country, state, city. & Same as $C_1$. \\ \hline
Payment & Pay application fee. & Not Applicable. \\ \hline
Education & Provide university, major and extra activities. & Same as $C_1$. \\ \hline
Profile Test & Online tests to measure profile characteristics such as ambition and interests. & Online tests to measure big 5 characteristics. \\ \hline
Experience & Write on professional experience using the model (S:situation, T:task, A:action, R:result). & Write about important professional experience \\ \hline
Logic Test & Perform online tests to map levels in problem-solving involving logic puzzles. & Same objective as $C_1$ but with specific test for $C_2$ \\ \hline
Video Submission & 2-min, explaining why they deserve the fellowship. & 5-min, making a case to be selected for the position. \\ \hline
Video Evaluation & Applicants are evaluated based on their entire profile submitted. & Same as $C_1$ but with different criteria. \\ \hline
Interview & 1-on-1 interview to clarify questions about the applicant's profile. & Same as $C_1$ but with different criteria. \\ \hline
Panel & Former fellows interview 5 to 6 applicants at the same time in a group discussion. & Managers interview 4 applicants in a group discussion. \\ \hline
Committee & Senior fellows and selection team select applicants to move to the final step. & Not Applicable. \\ \hline
Final & Applicants are interviewed by the board of the company. & Applicants are interviewed by a group of directors \\ \hline
\end{tabularx}
\end{table*}

Each stage in the process contains its own set of variables. For example, in the stage \textit{Demographics}, information about state and city is collected. Therefore, we refer to \textit{Demographics} features for those collected in the \textit{Demographics} stage. The data collected in each process is very similar in stages such as \textit{Demographics} and \textit{Education}, both in the form of content and structure. For stages with open questions such as \textit{Experience} and \textit{Video Submission}, each process has its own specific questions (See Table~\ref{tab:process_description} for details).
\subsection{Medical Diagnosis}
For medical diagnosis, we use two public datasets, Pima Indians Diabetes~\cite{bennett1971diabetes} and Thyroid Disease Data Set~\cite{quinlan1987inductive}. 
\subsubsection{Pima}
In the Pima Indians Diabetes dataset~\cite{bennett1971diabetes}, the goal is to predict if a patient has diabetes based on diagnostic measurements. These measurements come from questions and different lab tests with associated costs. For example, simple tests such as body mass index (BMI) and blood pressure cost 1 dollar. More elaborated tests have a higher cost, such as glucose blood test (17 dollars) and insulin test (23 dollars).

More expensive tests present more cases of missing data. Hence, we divide this process into two stages. Stage 1 contains 729 samples (34\% positive cases) with only the simpler features\footnote{Pregnancies, Blood Pressure, BMI, Diabetes Pedigree Function, Age}. Stage 2 contains 392 samples (32\% positive cases) with simpler and more expensive features\footnote{Skin Thickness,  Insulin, Glucose}.
\subsubsection{Thyroid}
The Thyroid dataset~\cite{quinlan1987inductive} consists of 21 clinical test results for a set of patients tested for thyroid dysfunction. Similar to Pima, the clinical tests also have different costs, and we use this information to define two stages. Stage 1 contains the tests that are easier/cheaper to obtain\footnote{Age, Sex, On Tryroxine, Query Thyroxine, On Antithyroid, Sick , Pregnant, Thyroid Surgery, I131 Treatment, Query Hypothyroid, Query Hyperthyroid, Lithium, Goitre, Tumour, Hypopituitary, Psych}, while stage 2 contains the more expensive tests\footnote{TSH, T3, TT4, T4U, FTI}.

Data from two years are available. For year 1, we have 3772 samples in stage 1 and 2752 in stage 2. For year 2, stage 1 has 3428, and stage 2 has 2534 samples. In all cases, the distribution of positive samples is around 7.5\%.
\subsection{Feature Preparation}
In all datasets, categorical variables are converted to numerical values using a standard one-hot encode transformation. 

For later stages in selection processes such as \textit{Video}, the speech is extracted, and the data is used as a text. To convert text data in \textit{Video} and \textit{Experience} to numerical values, we create word embeddings using Word2Vec~\cite{mikolov2013efficient}. The goal is to assign high-dimensional vectors (embeddings) to words in a text corpus while preserving their syntactic and semantic relationships. After obtaining embeddings for each world, we perform aggregation for a text (answer, video) using a simple average of the embedding vectors.
\subsection{Validation and Performance Metrics}
For selection processes and the Thyroid dataset, we perform longitudinal experiments, using a previous year as a training and test set and the following year as a validation set. For example, we split the dataset from $year^1$ in train and test, find the best model, and validate it using the dataset from $year^2$.

For selection processes, we also combine the datasets from $year^1$ and $year^2$ and validate the results in $year^3$, which results in 4 groups. For the train and test split, we also perform 10-fold cross-validation (CV), resulting in 40 runs for each company. For Thyroid, we have only 1 group (10 runs), since we can only train using $year^1$ and validate using $year^2$. For Pima, we perform 10-fold CV with the entire dataset (10 runs). 

We compare the models using F1-score for the positive class, which balances precision and recall for the selected applicants or diagnosed patients, in each stage.
\subsection{Settings and Benchmark Methods}
To compare our framework with established methods and study the effect of each component, we define four settings.
\subsubsection{AEMTD}
we first define a baseline model using \textit{Support Vector Machines} (SVM)~\cite{chang2011libsvm}. We tested other standard algorithms such as \textit{Logistic Regression} and standard NNs, but SVM obtained the best results. We train multiple binary classifiers (MBT), one for each stage, using SL to predict if the applicant is accepted, or patient diagnosed. This baseline is called N-MBT-SL and the classifier SVM-N. To evaluate the effect of AEMTD, we train the individual classifiers (SVM-C) again but with the complete dataset generated from AEMTD (AEMTD-MBT-SL).
\subsubsection{MLSSL}
In this setting, we train only one classifier using the complete dataset from AEMTD. For the labels, we design the problem as a multi-class classification so that the label for each observation is the last stage the individual was rejected or diagnosed. For example, an applicant rejected in stage \textit{Logic Test} has label 5, while an another approved in all stages has label 12. We then train an individual multi-class (IMC) classifier to predict each of these labels. 

We call this experiment AEMTD-IMC-SL and for the classifiers, we use SVM and a DNN method with \textit{Virtual Adversarial Training (VAT)}~\cite{miyato2018virtual} using $K=1, \epsilon=2$, $\alpha=1$.

In the final experiments, we create an individual multi-label (IML) classifier using SSL (AEMTD-IML-SSL). We compare our method with MLML~\cite{wu2014multi} and C2AE~\cite{yeh2017learning} using the parameter values defined in their respective papers. For STAGE, we use~$\lambda=0.5$, $k_{nn}=20$ and $h_{nn}=5$. For NN-based methods, we use dense NNs with fully connected layers, and the structure is chosen to have a similar number of parameters for all methods in each setting.

\section{Results}
\label{sec:results}
In this section, we present the results in all experiments.
\subsection{Selection processes}
The results for $C_1$ and $C_2$ are shown in Fig.~\ref{fig:c1_c2_results} and we analyze them considering two aspects:
\begin{figure*}[tbp]
\centerline{\includegraphics[width=\linewidth]{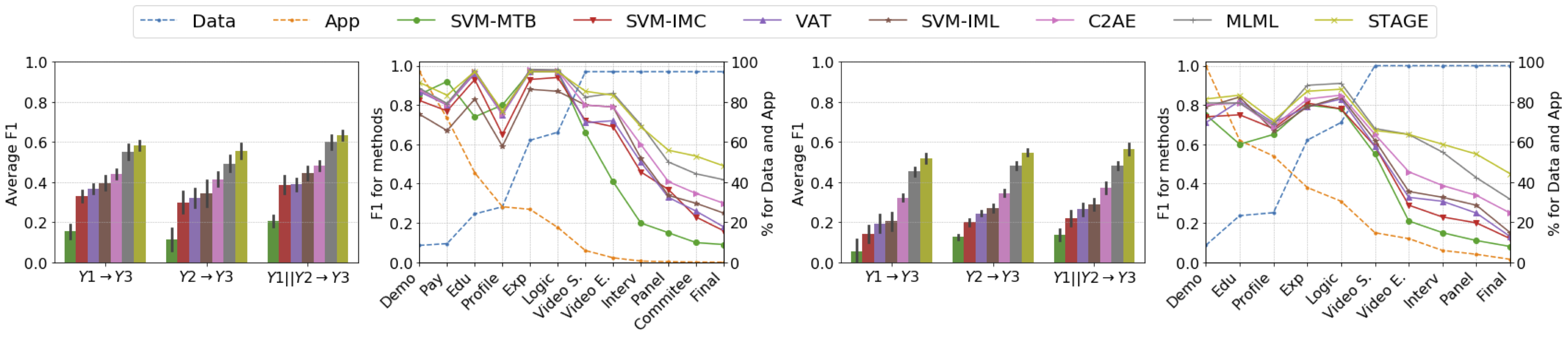}}
\caption{ LEFT - results for $C_1$. RIGHT - results for $C_2$. Y\textit{X} means $year^\textit{x}$. Bar-plots indicate average results for all years in later stages (Video E. to Final). Line-plots indicate results for a single year ($Y1||Y2\rightarrow Y3$) in all stages. In line-plots, for each new stage, the number of applicants drops while more data is obtained. Methods perform better with more samples, but the performance decreases substantially in the later stages. \textit{Data} refers to the number of features in each stage compared to the final dataset. \textit{App} refers to the number of applicants in each stage compared to the number in the first stage.}
\label{fig:c1_c2_results}
\end{figure*}
\subsubsection{General results across companies}
Comparing the first two settings (N-MBT-SL and AEMTD-MBT-SL), when individual classifiers are trained for each stage, we see that AEMTD is important as the performance of SVM-C is better than SVM-N in all experiments. This shows that the predictive future features help, even for individual classifiers trained in reduced datasets in later stages.

When the problem is redefined as multi-class (AEMTD-IMC-SL), and one classifier is trained for all stages, we see that all algorithms perform better than individual classifiers. We observe that SVM-IMC performs similar to VAT, which uses a regularized DNN. Both of the methods still suffer from overfitting in later stages as only a few samples have labels from that stage, and no information about label correlation is given to the classifiers. 

As expected, the best results are obtained in the last setting (AEMTD-IML-SSL), when MLSSL is used. The ML approach gives information about label correlation to the classifier, and the SSL component incorporates the relationships between labeled and unlabelled samples. We observe that MLML performs than C2AE. This is expected as the MLML method carefully designs the assumptions to guarantee label smoothness in the predictions, whereas C2AE expects the label embedding to self-learn these characteristics. The best performing method is our proposed STAGE. The additional label sequence constraint imposed to guarantee sequence consistency helps the classifier to achieve better performance in later stages. 
\subsubsection{Company Results}
\label{subsec:company_results}
Fig.~\ref{fig:c1_c2_results}-LEFT shows that methods are consistent for ${C_1}$ across years. We observe that predictions for $year^{3}$ using $year^{1}$ is better than using $year^{2}$. This suggests that the profile for applicants changes across years, and the ones approved in $year^{3}$ have more similarities with those approved in $year^{1}$. When using combined data ($year^{1}||year^{2}$), we achieve better results due to the increase in sample size and the combination of profiles from different years.

Results for $C_2$ are shown in Fig.~\ref{fig:c1_c2_results}-RIGHT. In this process, the number of applicants is more evenly distributed, and more applicants reach the final stages, which causes the average performance to be more similar across all stages. Differently from $C_1$, we see that it is better to use data from $year^2$ than $year^1$ to predict results in $year^3$. We also see the standard deviation being higher in the experiments using $year^1$.

In stages, it is clear from the line-plots that methods perform well while there is enough data to generalize. However, in later stages, there is a drop in performance caused by the small sample size. For both companies, algorithms based on AEMTD-IML-SSL perform significantly better with results around 3$\times$ higher than baseline (SVM-N) for later stages. Additionally, our method outperforms the second-best in later stages with a gain of 7\% and 14\% for $C_1$ and $C_2$, respectively.
\subsection{Medical Diagnosis}
Results F1 Score on positive class for the PIMA and Thyroid datasets are shown in Table~\ref{tab:results_medical}. In general, the performance for all methods is similar to what was observed in the experiments with selection processes. The algorithms in AEMTD-IML-SSL can outperform other methods, including training individual classifiers. However, for the PIMA dataset, the gain when using the best algorithms over the baseline is not as impressive as in other cases. The small sample size, in general, even in early stages, make more complex algorithms to overfit. For example, SVM-N and SVM-C are comparable to SVM-IMC and better than the DNN method with VAT. 

For the Thyroid dataset, the larger sample size in both stages helps complex models to achieve more significant gains. In this case, the challenge is the unbalanced distribution of the labels, which explains the relatively low results for the F1 score. However, our method STAGE achieved the best performance with a gain of almost 2$\times$ the baseline method. In general, STAGE outperforms or matches the second the best algorithms, which show the effectiveness of our approach.
\begin{table}[tbp]
\caption{Results in Medical Diagnosis}
\label{tab:results_medical}
\centering
\scriptsize
\begin{tabular}{|c|c|c|c|}
\hline
Settings                       & Methods & PIMA (CV)                 & Thyroid (LG)              \\ \hline
N-MBT-SL                       & SVM-N   & 0.25 $\pm$ 0.045          & 0.24 $\pm$ 0.052          \\ \hline
AEMTD-MBT-SL                   & SVM-C   & 0.27 $\pm$ 0.063          & 0.25 $\pm$ 0.066          \\ \hline
\multirow{2}{*}{AEMTD-IMC-SL}  & VAT     & 0.18 $\pm$ 0.074          & 0.36 $\pm$ 0.05           \\ \cline{2-4} 
                               & SVM-IMC & 0.31 $\pm$ 0.06           & 0.36 $\pm$ 0.056          \\ \hline
\multirow{3}{*}{AEMTD-IML-SSL} & C2AE    & 0.31 $\pm$ 0.054          & 0.43 $\pm$ 0.06           \\ \cline{2-4} 
                               & MLML    & \textbf{0.32 $\pm$ 0.077} & 0.43 $\pm$ 0.055          \\ \cline{2-4} 
                               & STAGE   & \textbf{0.32 $\pm$ 0.039} & \textbf{0.45 $\pm$ 0.054} \\ \hline
\end{tabular}
\end{table}
\section{Conclusion}
\label{sec:conclusion}
We presented a framework that combines AAE and MTL to create an adversarial-encoder-multi-task-decoder (AEMTD) structure. Given a dataset in a stage, our structure can learn the distribution of current features, reproduce them and predict future features generating a complete dataset. For classification, we combine ML and SSL so that an MLSSL classifier can learn the relationships between labels as well as use labeled and unlabeled examples during training. Additionally, we propose a sequence constraint to guarantee that the predictions are consistent with the sequence structure of a multi-stage process. We call this method STAGE, and we show its effectiveness by performing experiments using real-world data from two different domains (selection processes and medical diagnosis). Stage outperformed or matched all the other methods, achieving gains as high as 3$\times$ standard baselines and 14\% over other state-of-art algorithms. 

For future research, we would like to derive a method to connect the two components so that the classification information can be used during feature prediction in an end-to-end process. Additionally, selection processes are a sensitive topic, and it is possible that training on existing outcomes might reproduce inherent biases. Therefore, we would like to combine our method with fairness and equality constraints.

\bibliographystyle{IEEEtran}
\bibliography{main}

\end{document}